\documentclass[times,11pt]{article} 

\usepackage{amssymb} 
\usepackage{amsmath} 
\usepackage{latexsym} 
\usepackage{epsf} 
\usepackage[all]{xy}  
\usepackage{lscape} 
\usepackage{verbatim}    
\usepackage{times}  
\usepackage[margin=25mm]{geometry}

\mathchardef\mhyphen="2D

\title{\bf  Concrete Sentence Spaces for Compositional Distributional Models of Meaning}  

\author{Edward Grefenstette$^\ast$, Mehrnoosh Sadrzadeh$^\ast$, Stephen Clark$^\dagger$, Bob Coecke$^\ast$, Stephen Pulman$^\ast$\\ $^\ast$Oxford University Computing Laboratory, $^\dagger$University of Cambridge Computer Laboratory \\
 \texttt{\small firstname.lastname@comlab.ox.ac.uk}, \texttt{\small stephen.clark@cl.cam.ac.uk}}
\date{}  
 
\begin{document}  
\maketitle 

\begin{abstract}

Coecke, Sadrzadeh, and Clark \cite{LambekFest} developed a
compositional model of meaning for distributional semantics, in which
each word in a sentence has a meaning vector and the distributional
meaning of the sentence is a function of the tensor products of the
word vectors. Abstractly speaking, this function is the morphism
corresponding to the grammatical structure of the sentence in the
category of finite dimensional vector spaces. In this paper, we
provide a concrete method for implementing this linear meaning map, by
constructing a corpus-based vector space for the type of sentence. Our
construction method is based on structured vector spaces whereby
meaning vectors of all sentences, regardless of their grammatical
structure, live in the same vector space. Our proposed sentence space
is the tensor product of two noun spaces, in which the basis vectors
are pairs of words each augmented with a grammatical role. This
enables us to compare meanings of sentences by simply taking the inner
product of their vectors.

\end{abstract}

\thispagestyle{empty}

\section{Background}

Coecke, Sadrzadeh, and Clark \cite{LambekFest} develop a mathematical framework for a compositional
distributional model of meaning, based on the intuition
that \emph{syntactic analysis guides the semantic vector
composition}. The setting consists of two parts: a formalism for a
type-logical syntax and a formalism for vector space semantics. Each
word is assigned a grammatical type and a meaning vector in the space
corresponding to its type.  The meaning of a sentence is obtained by
applying the function corresponding to the grammatical structure of
the sentence to the tensor product of the meanings of the words in the
sentence. Based on the type-logic used, some words will have atomic
types and some compound function types. The compound types live in a
tensor space where the vectors are weighted sums (i.e. superpositions)
of the pairs of bases from each space. Compound types are ``applied''
to their arguments by taking inner products, in a similar manner to how
predicates are applied to their arguments in Montague semantics.

For the type-logic we use Lambek's Pregroup grammars
\cite{LambekBook}. The use of pregoups is not essential, but leads to
a more elegant formalism, given its proximity to the categorical
structure of vector spaces (see \cite{LambekFest}). A Pregroup is a
partially ordered monoid where each element has a right and left
cancelling element, referred to as an \emph{adjoint}. It can be seen
as the algebraic counterpart of the cancellation calculus of
Harris~\cite{Harris}. The operational difference between a Pregroup
and Lambek's Syntactic Calculus is that, in the latter, the monoid
multiplication of the algebra (used to model juxtaposition of the
types of the words) has a right and a left adjoint, whereas in the
pregroup it is the elements themselves which have adjoints. The
adjoint types are used to denote functions, e.g. that of a transitive
verb with a subject and object as input and a sentence as output. In
the Pregroup setting, these function types are still denoted by
adjoints, but this time the adjoints of the elements themselves.
 
As an example, consider the sentence ``dogs chase cats''. We assign
the type $n$ (for noun phrase) to ``dog'' and ``cat'', and $n^rsn^l$
to ``chase'', where $n^r$ and $n^l$ are the right and left adjoints of
$n$ and $s$ is the type of a (declarative) sentence. The
type $n^rsn^l$ expresses the fact that the verb is a predicate that
takes two arguments of type $n$ as input, on its right and left, and outputs
the type $s$ of a sentence. The parsing of the sentence is the
following reduction: \[ n (n^r s n^l) n\leq 1s1 = s \]
 This parse is based on the cancellation of $n$ and $n^r$, and also $n^l$ and $n$; i.e. $n n^r \leq 1$ and $n^l n \leq 1$ for 1 the unit of
juxtaposition. The reduction expresses the fact that the
juxtapositions of the types of the words reduce to the type of a
sentence.    

On the semantic side, we assign the vector space $N$ to the type $n$,
and the tensor space $N \otimes S \otimes N$ to the type $n^r s n^l$.
Very briefly, and in order to introduce some notation, recall that the
tensor space $A \otimes B$ has as a basis the cartesian product of a
basis of $A$ with a basis of $B$. Recall also that any vector can be
expressed as a weighted sum of basis vectors; e.g. if
$(\overrightarrow{v_1},\ldots,\overrightarrow{v_n})$ is a basis of
$A$ then any vector $\overrightarrow{a}\in A$ can be written as
$\overrightarrow{a} = \sum_i{C_i \overrightarrow{v_i}}$ where each
$C_i \in \mathbb{R}$ is a weighting factor. Now for
$(\overrightarrow{v_1},\ldots,\overrightarrow{v_n})$ a basis of $A$
and $(\overrightarrow{v'_1},\ldots,\overrightarrow{v'_n})$ a basis
of $B$, a vector $\overrightarrow{c}$ in the tensor space $A \otimes
B$ can be expressed as follows: \[ \sum_{ij}{C_{ij} \,
(\overrightarrow{v_i}\otimes\overrightarrow{v'_j})} \] where the
tensor of basis vectors
$\overrightarrow{v_i}\otimes\overrightarrow{v'_j}$ stands for their
pair $(\overrightarrow{v_i},\overrightarrow{v'_j})$. In general
$\overrightarrow{c}$ is not separable into the tensor of two vectors,
except for the case when $\overrightarrow{c}$ is not {\em entangled}. For non-entangled vectors we can write
$\overrightarrow{c} = \overrightarrow{a} \otimes \overrightarrow{b}$
for $\overrightarrow{a} = \sum_i{C_i \overrightarrow{v_i}}$ and
$\overrightarrow{b} = \sum_j{C'_j \overrightarrow{v'_j}}$; hence the
weighting factor of $\overrightarrow{c}$ can be obtained by simply
multiplying the weights of its tensored counterparts, i.e.  $C_{ij} =
C_i \times C'_j$. In the entangled case these weights cannot
be determined as such and range over all the possibilities. We take advantage of this fact  to encode meanings of verbs, and
in general all words that have compound types and are interpreted
as predicates, relations, or functions. For a brief discussion see the last paragraph of this section. Finally, we use the Dirac notation to denote  the dot or inner product of two vectors  $\langle \overrightarrow{a} \mid \overrightarrow{b} \rangle \in \mathbb{R}$ defined by $\sum_{i} C_i \times C'_i$.


Returning to our example, for the meanings of nouns we have
$\overrightarrow{\text{dogs}}, \overrightarrow{\text{cats}} \in N$,
and for the meanings of verbs we have $\overrightarrow{\text{chase}} \in
N \otimes S \otimes N$, i.e. the following superposition: \[
\sum_{ijk} C_{ijk} \,(\overrightarrow{n_i} \otimes
\overrightarrow{s_j} \otimes \overrightarrow{n_k}) \]
 Here $\overrightarrow{n_i}$ and $\overrightarrow{n_k}$ are basis
 vectors of $N$ and $\overrightarrow{s_j}$ is a basis vector of
 $S$. From the categorical translation method presented
 in~\cite{LambekFest} and the grammatical reduction $n (n^r s n^l)
 n\leq s$, we obtain the following linear map as the categorical
 morphism corresponding to the reduction: \[ \epsilon_N \otimes 1_s
 \otimes \epsilon_N : N \otimes (N \otimes S \otimes N) \otimes N \to
 S \]
 Using this map, the
meaning of the sentence is computed as follows:
 \begin{align*}
 \overrightarrow{\text{dogs} \ \text{chase} \ \text{cats}} &\quad = \quad \left(\epsilon_N \otimes 1_s \otimes \epsilon_N\right) \left(\overrightarrow{\text{dogs}} \otimes \overrightarrow{\text{chase}} \otimes \overrightarrow{\text{cats}}\right)\\
 & \quad = \quad \left(\epsilon_N \otimes 1_s \otimes \epsilon_N\right) \left(\overrightarrow{\text{dogs}} \otimes \left(\sum_{ijk} C_{ijk} (\overrightarrow{n_i} \otimes \overrightarrow{s_j} \otimes \overrightarrow{n_k})\right) \otimes \overrightarrow{\text{cats}} \right)\\
 &\quad = \quad \sum_{ijk} C_{ijk} \langle \overrightarrow{\text{dogs}} \mid \overrightarrow{n_i}\rangle  \overrightarrow{s_j} \langle \overrightarrow{n_k}\mid \overrightarrow{\text{cats}}\rangle\vspace{-2mm}
\end{align*}

The key features of this operation are, first, that the inner-products
reduce dimensionality by `consuming' tensored vectors and by virtue of
the following component function: \[ \epsilon_N : N \otimes N \to
\mathbb{R} :: \overrightarrow{a} \otimes \overrightarrow{b} \mapsto
\langle \overrightarrow{a} \mid \overrightarrow{b} \rangle \]
 Thus the tensored word vectors $\overrightarrow{\text{dogs}} \otimes
 \overrightarrow{\text{chase}} \otimes \overrightarrow{\text{cats}}$
 are mapped into a sentence space $S$ which is common to all sentences
 regardless of their grammatical structure or
 complexity. Second, note that the tensor product
 $\overrightarrow{\text{dogs}} \otimes \overrightarrow{\text{chase}}
 \otimes \overrightarrow{\text{cats}}$ does not need to be calculated,
 since all that is required for computation of the sentence vector are
 the noun vectors and the $C_{ijk}$ weights
 for the verb. Note also that the inner product operations are simply
 picking
 out basis vectors in the noun space, an
 operation that can be performed in constant time. Hence this
 formalism avoids 
 two problems faced by approaches in the vein of
 ~\cite{Smolensky,ClarkPulman}, which use the tensor product as a
 composition operation: first, that the sentence meaning space is high
 dimensional and grammatically different sentences have
 representations with different dimensionalities, preventing them from being
 compared directly using inner products; and second, that the space
 complexity of the tensored representation grows exponentially with
 the length and grammatical complexity of the sentence. In constrast, the
 model we propose does not require the tensored vectors being combined
 to be represented explicitly.

Note that  we have taken the vector of the transitive verb, e.g. $\overrightarrow{\text{chase}}$, to be an entangled vector in the tensor space $N \otimes S \otimes N$. But why can this not be a separable vector, in which case the meaning of the verb would be as follows:
\[
\overrightarrow{\text{chase}} \quad = \quad \sum_{i}{C_i \overrightarrow{n_i}} \quad \otimes \quad \sum_{j}{C'_j \overrightarrow{s_j}} \quad \otimes \quad \sum_{k}{C''_k \overrightarrow{n_k}}
\]
The meaning of the sentence would then become  $\sigma_1 \sigma_2 \sum_{j}{C'_j \overrightarrow{s_j}}$ for $\sigma _1 = \sum_{i}{C_i \langle \overrightarrow{\text{dogs}} \mid\overrightarrow{n_i}\rangle}$ and $\sigma_2 = \sum_{k}{C''_k \langle \overrightarrow{\text{cats}}\mid\overrightarrow{n_k}\rangle}$. The problem is that this meaning only depends on the meaning of the verb and is  independent of the meanings of the subject and object, whereas the meaning from the entangled case, i.e.~$\sigma_1 \sigma_2\sum_{ijk}{C_{ijk} \overrightarrow{s_j}}$, 
depends on the meanings of subject and object as well as the verb. 

\section{From Truth-Theoretic to Corpus-based Meaning}

The model presented above is compositional and distributional, but still abstract. To make it concrete,  $N$ and $S$ have to be constructed by providing a method for determining the $C_{ijk}$ weightings. 
Coecke, Sadrzadeh, and Clark \cite{LambekFest} show how a  truth-theoretic meaning can be derived in the compositional framework. For example,
 assume that $N$ is spanned by all animals and $S$ is the two-dimensional space spanned by $\overrightarrow{\text{true}}$ and $\overrightarrow{\text{false}}$. We use the weighting factor to define a model-theoretic meaning for the verb as follows:
\[
C_{ijk} \overrightarrow{s_j} = \begin{cases} \overrightarrow{\text{true}} &
{chase} (\overrightarrow{n_i}, \overrightarrow{n_k})  =  \text{true}\,,\\  \overrightarrow{\text{false}} & o.w.\end{cases}
\]
The definition of our meaning map ensures that this value propagates
to the meaning of the whole sentence. So $chase(\overrightarrow{dogs}, \overrightarrow{cats})$ becomes true whenever ``dogs chase cats'' is true and false otherwise. This is exactly how meaning is computed in the model-theoretic view on semantics.  One way to generalise this
truth-theoretic meaning is to assume that ${chase}
(\overrightarrow{n_i}, \overrightarrow{n_k}) $ has degrees of truth, for instance by defining $chase$ as a combination of $run$ and $catch$, such as:
\[
chase = {2 \over 3}  run + {1 \over 3}  catch
\]
 Again, the meaning map ensures that these
degrees propagate to the meaning of the whole sentence. For a worked out example see~\cite{LambekFest}. But neither of
these examples provide a {\em distributional} sentence meaning. 

Here we  take a first
step towards a corpus-based distributional model, by attempting to recover a  meaning for a sentence based on the meanings of the words derived from a corpus. But crucially this meaning goes beyond just composing the meanings of words using a vector operator, such as tensor product, summation or multiplication \cite{Lapata}. Our computation of sentence meaning  treats some vectors as functions and others as function arguments, according to how the words in the sentence are typed, and uses the syntactic structure as a guide to determine how the functions are applied to their arguments. The intuition behind this approach is that \emph{syntactic analysis guides semantic vector composition}. 



The contribution of this
paper is to introduce some concrete constructions for a compositional distributional model of
meaning. These constructions demonstrate how the
mathematical model of~\cite{LambekFest} can be implemented in a
concrete setting which introduces a richer, not necessarily truth-theoretic, notion of natural language semantics which is closer to the ideas underlying standard distributional models of word meaning. 
We leave full evaluation to future work, in order to determine whether 
the following method in conjunction with word vectors built from large corpora leads to improved results on 
 language
processing tasks, such as  computing sentence similarity and paraphrase evaluation.


\bigskip
\noindent
{\bf Nouns and Transitive Verbs.}
 We take  $N$ to be  a \emph{structured vector
space}, as in~\cite{ErkPado,Greffen}. The bases of $N$ are
annotated by 
`properties' obtained by combining dependency relations with nouns, verbs and adjectives. 
For example, basis vectors might be associated with properties such as ``arg-fluffy'', denoting the argument of the adjective fluffy, ``subj-chase'' denoting the subject of the verb chase, ``obj-buy'' denoting the object of the verb buy, and so on. We construct the vector for a noun by counting how many times in the corpus  a word has been the argument of `fluffy', the subject of `chase', the object of `buy', and so on. 

The framework in \cite{LambekFest} offers no guidance as to what the sentence space should consist of. Here we take the sentence space $S$ to be  $N \otimes N$,  so
its bases are of the form $\overrightarrow{s_j} =
{(\overrightarrow{n_i}, \overrightarrow {n_k})}$.  The intuition is that, for a transitive verb, the meaning of a sentence is determined by the meaning of the verb together with its subject and object.\footnote{Intransitive and ditransitive verbs are interpreted in an analagous fashion; see $\S$\ref{ssec:different_grammatical_structures}. }
The verb vectors
$C_{ijk}{(\overrightarrow{n_i}, \overrightarrow
{n_k})}$ are built by  counting how
many times a word that is  $n_i$ (e.g. has the property of being fluffy) has been subject of the verb and
a word that is $n_k$ (e.g. has the property that it's bought) has been its object, where the counts are moderated by the extent to which the subject and object exemplify each property (e.g. \emph{how fluffy} the subject is). To give a rough paraphrase of the intuition behind this approach, the meaning of ``dog chases cat" is given by: the extent to which a dog is fluffy and a cat is something that is bought (for the $N \otimes N$ property pair ``arg-fluffy" and ``obj-buy"), and the extent to which fluffy things {\em chase} things that are bought (accounting for the meaning of the verb for this particular property pair); plus the extent to which  a dog is something that runs and a cat is something that is cute (for the $N \otimes N$ pair ``subj-run" and ``arg-cute"), and the extent to which things that run {\em chase} things that are cute (accounting for the meaning of the verb for this particular property pair); and so on for all noun property pairs.

\bigskip
\noindent
{\bf Adjective Phrases.} Adjectives are dealt with in
a similar way. We give them the syntactic type $nn^l$ and
build their vectors in $N \otimes N$. The syntactic reduction $nn^l n
\to n$ associated with applying an adjective to a noun gives us the
map $1_N \otimes \epsilon_N$ by which we semantically compose an
adjective with a noun, as follows: \begin{displaymath}
	\overrightarrow{\text{red fox}} = (1_N \otimes \epsilon_N)(\overrightarrow{\text{red}} \otimes \overrightarrow{\text{fox}}) = \sum_{ij}{C_{ij}\overrightarrow{n_i} \langle \overrightarrow{n_j} \mid \overrightarrow{\text{fox}} \rangle}
\end{displaymath}
We can view the $C_{ij}$ counts as determining what sorts of properties the arguments of a particular adjective
typically have (e.g. arg-red, arg-colourful for the adjective
``red'').

\bigskip
\noindent
{\bf Prepositional Phrases.}  We assign the type $n^r n$ to the whole  prepositional phrase (when it modifies a noun), for example to ``in the forest'' in the sentence ``dogs chase cats in the forest''. The pregroup parsing is as follows:
\[
n (n^rsn^l) n (n^r n) \leq 1sn^l 1 n \leq sn^l n \leq s1 = s
\]
The vector space corresponding to the prepositional phrase will thus be the tensor space $N \otimes N$ and the categorification of the parse  will be the composition of two morphisms: $(1_S \otimes \epsilon^l_N) \circ (\epsilon^r_N \otimes 1_S \otimes 1_N \otimes \epsilon^r_N \otimes 1_N)$. The substitution specific to the prepositional phrase happens when computing the vector for ``cats in the forest'' as follows:
\begin{align*}
\overrightarrow{\text{cats \ in \ the \ forest}}  &= 
(\epsilon^r_N \otimes 1_N) \left (\overrightarrow{\text{cats}} \otimes \overrightarrow{\text{in \ the \ forest}}\right)\\
&= (\epsilon^r_N \otimes 1_N) \left (\overrightarrow{\text{cats}} \otimes \sum_{lw} C_{lw} \overrightarrow{n_l} \otimes \overrightarrow{n_k}\right)\\
& = \sum_{lw} C_{lw} \langle \overrightarrow{\text{cats}} \mid \overrightarrow{n_l} \rangle \overrightarrow{n_w}
\end{align*}
Here we set the weights $C_{lw}$ in a similar manner to the cases of adjective phrases and verbs with the counts determining what sorts of properties the noun modified by the prepositional phrase has, e.g.  the number of times something that has attribute $n_l$ has been in the forest. 

\bigskip
\noindent
{\bf Adverbs.} We assign the type $s^r s$ to the adverb, for example to ``quickly'' in the sentence ``Dogs chase cats quickly''. The pregroup parsing is as follows:
\[
n (n^rsn^l) n (s^rs) \leq 1s1s^rs = ss^rs\leq 1s = s
\]
 Its categorification will be a composition of two morphisms $(\epsilon^r_S \otimes 1_S) \circ (\epsilon^r_N \otimes 1_S \otimes \epsilon^l_N \otimes 1_S \otimes 1_S)$. The substitution specific to the adverb happens after computing the meaning of the sentence without it, i.e. that of ``Dogs chase cats'', and is as follows:
\begin{align*}
\overrightarrow{\text{Dogs \ chase \ cats \ quickly}} &= 
(\epsilon^r_S \otimes 1_S) \circ (\epsilon^r_N \otimes 1_S \otimes \epsilon^l_N \otimes 1_S \otimes 1_S) 
\left (\overrightarrow{\text{Dogs}} \otimes \overrightarrow{\text{chase}} \otimes  \overrightarrow{\text{cats}} \otimes \overrightarrow{\text{quickly}}\right)\\
&= 
(\epsilon^r_S \otimes 1_S)  \left (\sum_{ijk} C_{ijk} \langle \overrightarrow{\text{dogs}} \mid \overrightarrow{n_i}\rangle  \overrightarrow{s_j} \langle \overrightarrow{n_k}\mid \overrightarrow{\text{cats}}\rangle \otimes \overrightarrow{\text{quickly}}\right)\\
&= (\epsilon^r_S \otimes 1_S)  \left(\sum_{ijk} C_{ijk} \langle \overrightarrow{\text{dogs}} \mid \overrightarrow{n_i}\rangle  \overrightarrow{s_j} \langle \overrightarrow{n_k}\mid \overrightarrow{\text{cats}}\rangle  \otimes \sum_{lw} C_{lw} \overrightarrow{s_l} \otimes \overrightarrow{s_w}\right)\\
&= \sum_{lw} C_{lw} \left \langle \sum_{ijk} C_{ijk} \langle \overrightarrow{\text{dogs}} \mid \overrightarrow{n_i}\rangle  \overrightarrow{s_j} \langle \overrightarrow{n_k}\mid \overrightarrow{\text{cats}}\rangle \mid  \overrightarrow{s_l}\right \rangle 
\overrightarrow{s_k}
\end{align*}
The  $C_{lw}$ weights are defined in a similar manner to the above cases, i.e. according to the properties the adverb has, e.g. which verbs it has  modified. Note that now the basis vectors $\overrightarrow{s_l}$ and $\overrightarrow{s_w}$ are themselves pairs of basis vectors from the noun space, $(\overrightarrow{n_i}, \overrightarrow{n_j})$. Hence,  $C_{lw}(\overrightarrow{n_i}, \overrightarrow{n_j})$ can be set only for the case when $l=i$ and $w=j$; these counts determine what sorts of properties the verbs that happen quickly have (or more specifically what properties the subjects and objects of such verbs have). By taking the whole sentence into account in the interpretation of the adverb, we are in a better position to  semantically distinguish between the meaning of adverbs such as ``slowly'' and ``quickly'', for instance in terms of the properties that the verb's subjects have. For example, it is possible that elephants are more likely to be the subject of a verb which is happening slowly, e.g. run slowly, and cheetahs are more likely to be the subject of a verb which is happening quickly.




\section{Concrete Computations}

In this section we first describe how to obtain the relevant counts
from a parsed corpus, and then give some similarity calculations for
some example sentence pairs. 

Let $\mathcal{C}_l$ be the set of grammatical relations (GRs) for
sentence $s_l$ in the corpus. Define $\mathit{verbs}(\mathcal{C}_l)$ to be the
function which returns all instances of verbs in $\mathcal{C}_l$, and
$\mathit{subj}$ (and similarly $\mathit{obj}$) to be the function
which returns the subject of an instance $V_{\textit{instance}}$ of a verb $V$, for a particular set of GRs for a sentence:

\begin{displaymath}
	subj(V_{\textit{instance}}) = \begin{cases} noun & \text{if $V_{\textit{instance}}$ is a verb with subject $noun$}\\
	\varepsilon_n & o.w.\end{cases}
\end{displaymath}
where $\varepsilon_n$ is the empty string. We express $C_{ijk}$ for a verb $V$ as follows:
\begin{displaymath}
C_{ijk} = \begin{cases} \sum_l\sum_{v \in \mathit{verbs}(\mathcal{C}_l)}{\delta(v,V) \langle \overrightarrow{subj(v)} \mid \overrightarrow{n_i} \rangle \langle \overrightarrow{obj(v)} \mid \overrightarrow{n_k} \rangle} & \text{if}\ \overrightarrow{s_j} ={(\overrightarrow{n_i} , \overrightarrow{n_k})} \\
0 & o.w.\end{cases}\vspace{-3mm}
\end{displaymath}
where  $\delta(v,V) = 1$ \ if $v = V$ and 0 otherwise.
Thus we construct $C_{ijk}$ for verb $V$ only for cases where the
subject property $n_i$ and the object property $n_k$ are paired in the
basis $\overrightarrow{s_j}$. This is done by counting the number of
times the subject of $V$ has property $n_i$ and the object of $V$ has property $n_k$, then multiplying them, as
prescribed by the inner products (which simply pick out the properties $n_i$ and $n_k$ from the noun vectors for the subjects and objects). 





The procedure for calculating the verb vectors, based on the formulation above, is as follows:

\begin{enumerate}
	\item For each GR in a sentence, if the relation is $subject$ and the head is a verb, then find the complementary GR with $object$ as a relation and the same head verb. If none, set the object to $\varepsilon_n$.
	\item Retrieve the noun vectors $\overrightarrow{subject}, \overrightarrow{object}$ for the subject dependent and object dependent from previously constructed noun vectors.
	\item For each $(n_i,n_k) \in basis(N) \times basis(N)$ compute the inner-product of $\overrightarrow{n_i}$ with $\overrightarrow{subject}$ and $\overrightarrow{n_k}$ with $\overrightarrow{object}$ (which involves simply picking out the relevant basis vectors from the noun vectors). Multiply the inner-products and add this to $C_{ijk}$ for the verb, with $j$ such that $\overrightarrow{s_j} = (\overrightarrow{n_i},\overrightarrow{n_k})$.
\end{enumerate}

\noindent
The procedure for other grammatical types is similar, based on the definitions of $C$ weights for the semantics of these types.

We now give a number of example calculations. We first manually define the distributions for nouns, which in practice would be obtained from a corpus:
\begin{center}
	\begin{tabular}{l|ccccc}
\hline
		& bankers & cats & dogs & stock & kittens\\
		\hline\hline
		1. arg-fluffy 		& 0 & 7 & 3 & 0 & 2\\
		2. arg-ferocious 	& 4 & 1 & 6 & 0 & 0\\
		3. obj-buys 		& 0 & 4 & 2 & 7 & 0\\
		4. arg-shrewd 		& 6 & 3 & 1 & 0 & 1\\
		5. arg-valuable 	& 0 & 1 & 2 & 8 & 0\\
		\hline
	\end{tabular}
\end{center}
We aim to make these counts match our intuitions, in that bankers are shrewd and a little ferocious but not furry, cats are furry but not typically valuable, and so on.

We also define the distributions for the transitive verbs `chase', `pursue' and `sell', again manually specified according to our intuitions about how these verbs are used. Since in the formalism proposed above, $C_{ijk}=0$ if $\overrightarrow{s_j} \neq (\overrightarrow{n_i},\overrightarrow{n_k})$, we can simplify the weight matrices for transitive verbs to two dimensional $C_{ik}$ matrices as shown below, where $C_{ik}$ corresponds to the number of times the verb has a subject with attribute $n_i$ and an object with attribute $n_k$. For example, the matrix below encodes the fact that something ferocious ($i=2$) chases something fluffy ($k=1$) seven times in the hypothetical corpus from which we might have obtained these distributions.
\begin{displaymath}
	C^{\textrm{chase}} = \left[
	\begin{tabular}{ccccc}
	1 & 0 & 0 & 0 & 0 \\
	7 & 1 & 2 & 3 & 1 \\
	0 & 0 & 0 & 0 & 0   \\
	2 & 0 & 1 & 0 & 1  \\
	1 & 0 & 0 & 0 & 0   \\
	\end{tabular}
	\right]
	\quad
	C^{\textrm{pursue}} = \left[
	\begin{tabular}{ccccc}
	0 & 0 & 0 & 0 & 0 \\
	4 & 2 & 2 & 2 & 4 \\
	0 & 0 & 0 & 0 & 0   \\
	3 & 0 & 2 & 0 & 1  \\
	0 & 0 & 0 & 0 & 0   \\
	\end{tabular}
	\right]
	\quad
	C^{\textrm{sell}} = \left[
	\begin{tabular}{ccccc}
	0 & 0 & 0 & 0 & 0   \\
	0 & 0 & 3 & 0 & 4   \\
	0 & 0 & 0 & 0 & 0   \\
	0 & 0 & 5 & 0 & 8  \\
	0 & 0 & 1 & 0 & 1  \\
	\end{tabular}
	\right]
\end{displaymath}
These matrices can be used to perform sentence comparisons:
\begin{align*}
	\langle \overrightarrow{\text{dogs chase cats}} &  \mid \overrightarrow{\text{dogs pursue kittens}} \rangle  = \\
	 & = \left\langle \left(\sum_{ijk}{C^{\text{chase}}_{ijk} \langle \overrightarrow{\text{dogs}} \mid \overrightarrow{n_i}\rangle  \overrightarrow{s_j} \langle \overrightarrow{n_k}\mid \overrightarrow{\text{cats}}\rangle}\right) \right| \left. \left(\sum_{ijk}{C^{\text{pursue}}_{ijk} \langle \overrightarrow{\text{dogs}} \mid \overrightarrow{n_i}\rangle  \overrightarrow{s_j} \langle \overrightarrow{n_k}\mid \overrightarrow{\text{kittens}}\rangle} \right) \right\rangle \\
	 &\\
     & = \sum_{ijk}{C^{\text{chase}}_{ijk}C^{\text{pursue}}_{ijk} \langle \overrightarrow{\text{dogs}} \mid \overrightarrow{n_i}\rangle \langle \overrightarrow{\text{dogs}} \mid \overrightarrow{n_i}\rangle \langle \overrightarrow{n_k}\mid \overrightarrow{\text{cats}}\rangle \langle \overrightarrow{n_k}\mid \overrightarrow{\text{kittens}}\rangle}
\end{align*}
The raw number obtained from the above calculation is 14844. Normalising it  by the product of the length of both sentence vectors gives the cosine  value of $0.979$.  

Consider now the sentence comparison $\langle \overrightarrow{\text{dogs chase cats}}   \mid \overrightarrow{\text{cats chase dogs}} \rangle $. The sentences in this pair contain the same words but the different word orders give the sentences very different meanings.
The raw number calculated from this inner product is 7341, and its normalised cosine measure is $0.656$, which demonstrates the sharp drop in similarity obtained from changing sentence structure. We expect some similarity since there is some non-trivial overlap between the properties identifying cats and those identifying dogs (namely those salient to the act of chasing). 

Our final example for transitive sentences is $\langle \overrightarrow{\text{dogs chase cats}}   \mid \overrightarrow{\text{bankers sell stock}} \rangle $, as two sentences that diverge in meaning completely.
The raw number for this inner product is 6024, and its cosine measure is $0.042$, demonstrating the very low semantic similarity between these two sentences. 



Next we consider some examples involving adjective-noun modification.
The $C_{ij}$ counts for an adjective $A$ are obtained in a similar manner to transitive or intransitive verbs:
\begin{displaymath}
C_{ij} = \begin{cases} \sum_l\sum_{a\in\mathit{adjs}(\mathcal{C}_l)}{\delta(a,A) \langle \overrightarrow{arg\mhyphen{}of(a)} \mid \overrightarrow{n_i} \rangle } & \text{if}\ \overrightarrow{n_i} = \overrightarrow{n_j} \\
0 & o.w.\end{cases}
\end{displaymath}
where $\mathit{adjs}(\mathcal{C}_l)$ returns all instances of adjectives in $\mathcal{C}_l$;
$\delta(a,A) = 1$ if $a = A$ and $0$ otherwise; and
$arg\mhyphen{}of(a) = noun$ if $a$ is an adjective with argument
$noun$, and $\varepsilon_n$ otherwise.

As before, we stipulate the $C_{ij}$ matrices by hand (and we eliminate all cases where $i \neq j$ since $C_{ij} = 0$ by definition in such cases):
\[
C^{\text{fluffy}}  = [9\ 3\ 4\ 2\ 2] \qquad
C^{\text{shrewd}}  = [0\ 3\ 1\ 9\ 1]\qquad 
C^{\text{valuable}}  = [3\ 0\ 8\ 1\ 8]
\]
We compute vectors for ``fluffy dog'' and ``shrewd banker'' as follows:
{\small \begin{align*}
\overrightarrow{\text{fluffy \ dog}} &= (3 \cdot 9)\, \overrightarrow{\text{arg-fluffy}} + (6 \cdot 3)\, \overrightarrow{\text{arg-ferocious}} + (2 \cdot 4)\, \overrightarrow{\text{obj-buys}} + (5 \cdot 2)\, \overrightarrow{\text{arg-shrewd}} + (2 \cdot 2)\, \overrightarrow{\text{arg-valuable}}\\
\overrightarrow{\text{shrewd \ banker}} &=( 0 \cdot 0) \, \overrightarrow{\text{arg-fluffy}} + (4 \cdot 3)\, \overrightarrow{\text{arg-ferocious}} + (0 \cdot 0)\, \overrightarrow{\text{obj-buys}} + (6 \cdot 9)\, \overrightarrow{\text{arg-shrewd}} + (0 \cdot 1)\, \overrightarrow{\text{arg-valuable}}
\end{align*}}
Vectors for $\overrightarrow{\text{fluffy \ cat}}$ and $\overrightarrow{\text{valuable \ stock}}$ are computed similarly. We obtain the following similarity measures:
\[
cosine ( \overrightarrow{\text{fluffy \ dog}} ,  \overrightarrow{\text{shrewd \ banker}} ) = 0.389\qquad
cosine ( \overrightarrow{\text{fluffy \ cat}} ,  \overrightarrow{\text{valuable \ stock}}) = 0.184
\]
These calculations carry over to sentences which contain the adjective-noun pairings compositionally and we obtain an  even lower similarity measure between sentences:
\[cosine( \overrightarrow{\text{fluffy dogs chase fluffy cats}} , \overrightarrow{\text{shrewd bankers sell valuable stock}} ) = 0.016
\]

\medskip
\noindent
To summarise, our example vectors provide us with the following similarity measures:
\begin{center}
\begin{tabular}{l|l|c}
\hline

{\bf Sentence 1} & {\bf Sentence 2} & {\bf Degree of similarity}\\
\hline\hline

dogs chase cats &dogs pursue kittens & $0.979$\\

dogs chase cats & cats chase dogs& $0.656$\\

dogs chase cats & bankers sell stock& $0.042$\\

fluffy dogs chase fluffy cats & shrewd bankers sell valuable stock & $0.016$\\
\hline
\end{tabular}\end{center}

\section{Different Grammatical Structures}
\label{ssec:different_grammatical_structures}

So far we have only presented the treatment of sentences with
transitive verbs. For sentences with intransitive verbs, the sentence
space suffices to be just $N$.  To compare the meaning of a transitive
sentence with an intransitive one, we embed the meaning of the latter
from $N$ into the former $N \otimes N$, by taking
$\overrightarrow{\varepsilon_n}$ (the `object' of an intransitive verb) to be $\sum_i{\overrightarrow{n_i}}$,
i.e. the superposition of all basis vectors of $N$. 

Following the method for the transitive verb, we calculate $C_{ijk}$ for an
instransitive verb $V$ and basis pair $\overrightarrow{s_j} =
{(\overrightarrow{n_i} , \overrightarrow{n_k})}$  as follows, where $l$ ranges over the sentences in the corpus:
\[ \sum_l\sum_{v \in \mathit{verbs}(\mathcal{C}_l)}{\delta(v,V)
\langle \overrightarrow{subj(v)} \mid
\overrightarrow{n_i} \rangle \langle
\overrightarrow{obj(v)} \mid \overrightarrow{n_k}
\rangle}
	 = \sum_l\sum_{v \in \mathit{verbs}(\mathcal{C}_l)}{\delta(v,V) \langle \overrightarrow{subj(v)} \mid \overrightarrow{n_i} \rangle \langle \overrightarrow{\varepsilon_n} \mid \overrightarrow{n_k} \rangle}
\]
and $\langle
\overrightarrow{\varepsilon_n} \mid \overrightarrow{n_i} \rangle = 1$
for any basis vector $n_i$.

We can now compare the meanings of transitive and intransitive
sentences by taking the inner product of their meanings
(despite the different arities of the verbs) and then normalising it by vector
length to obtain the cosine measure. For example:

\begin{align*}
\langle \overrightarrow{\text{dogs chase cats}} \mid \overrightarrow{\text{dogs chase}} \rangle & = \left\langle \left(\sum_{ijk}{C_{ijk} \langle \overrightarrow{\text{dogs}} \mid \overrightarrow{n_i}\rangle  \overrightarrow{s_j} \langle \overrightarrow{n_k}\mid \overrightarrow{\text{cats
}}\rangle}\right) \right| \left. \left(\sum_{ijk}{C'_{ijk} \langle \overrightarrow{\text{dogs}} \mid \overrightarrow{n_i}\rangle  \overrightarrow{s_j}} \right) \right\rangle\\
&\\
& = \sum_{ijk}{C_{ijk}C'_{ijk} \langle \overrightarrow{\text{dogs}} 
\mid \overrightarrow{n_i}\rangle \langle \overrightarrow{\text{dogs}} \mid 
\overrightarrow{n_i}\rangle \langle \overrightarrow{n_k}\mid \overrightarrow{\text{cats}}\rangle} 
\end{align*}
The raw number for the inner product is 14092  and its normalised cosine measure  is 0.961, indicating high similarity (but some difference) between a sentence with a transitive verb and one where the subject remains the same, but the verb is used intransitively.

\bigskip
\noindent
Comparing sentences containing nouns modified by adjectives to sentences with unmodified nouns is straightforward:
\begin{align*}
	\langle \overrightarrow{\textrm{fluffy dogs chase fluffy cats}} \mid \overrightarrow{\textrm{dogs chase cats}}\rangle &=\\
	&\\
	\sum_{ij} C^{\text{fluffy}}_i C^{\text{fluffy}}_j C^{\text{chase}}_{ij} C^{\text{chase}}_{ij} \langle \overrightarrow{dogs} \mid \overrightarrow{n_i} \rangle^2 \langle \overrightarrow{n_j} \mid \overrightarrow{cats} \rangle^2 &= 2437005\\	
\end{align*}
From the above we obtain the following similarity measure:
\[
cosine(\overrightarrow{\textrm{fluffy dogs chase fluffy cats}},\overrightarrow{\textrm{dogs chase cats}}) = 0.971
\]
For sentences with ditransitive verbs, the sentence space changes to $N \otimes N \otimes N$, on the basis of  the verb needing two objects; hence its grammatical type changes to $n^r s n^l n^l$. The transitive and intransitive verbs are embedded in this larger space in a similar manner to that described above; hence comparison of their meanings becomes possible.

\section{Ambiguous Words}
The two different meanings of a word can be distinguished by the different properties that they have. These properties are reflected in the corpus, by the different contexts in which the words appear. Consider the following example from~\cite{ErkPado}: the verb ``catch'' has two different meanings, ``grab'' and ``contract''. They are  reflected in the two sentences ``catch a ball'' and ``catch a disease''.
The compositional feature of our meaning computation enables us to realise the different properties of  the context words via the grammatical roles they take in the corpus. For instance, the word `ball' occurs as argument of `round', and so has a high weight for the base `arg-round', whereas the word `disease'  has a high weight  for the base `arg-contagious' and as `mod-of-heart'. We extend our example corpus from previously to reflect these differences as follows:
\begin{center}
	\begin{tabular}{l|cc}
		\hline
		& ball & disease\\
		\hline\hline
		1. arg-fluffy 		&1&0\\
		2. arg-ferocious 	&0&0\\
		3. obj-buys 		&5&0\\
		4. arg-shrewd 		&0&0\\
		5. arg-valuable 	&1&0\\
		6. arg-round 		&8&0\\
		7. arg-contagious 	&0&7\\
		8. mod-of-heart 	&0&6\\
		\hline
	\end{tabular}
\end{center}
In a similar way, we build a matrix for the verb `catch' as follows:
\[
C^{\textrm{catch}} = \left[
	\begin{tabular}{cccccccc}
		3 & 2 & 3 & 3 & 3 & 8 & 6 & 2\\
		3 & 2 & 3 & 0 & 1 & 4 & 7 & 4\\
		2 & 4 & 7 & 1 & 1 & 6 & 2 & 2\\
		3 & 1 & 2 & 0 & 0 & 3 & 6 & 2\\
		1 & 1 & 1 & 0 & 0 & 2 & 0 & 1\\
		0 & 0 & 0 & 0 & 0 & 0 & 0 & 0\\
		0 & 0 & 0 & 0 & 0 & 0 & 0 & 0\\
		0 & 0 & 0 & 0 & 0 & 0 & 0 & 0\\
	\end{tabular}
	\right]
\]
The last three rows are zero because we have assumed that the words that can take these roles are mostly objects and hence cannot catch anything.  Given these values, we compute the similarity measure between the two sentences ``dogs catch a ball'' and ``dogs catch a disease'' as follows:
\[
\langle \overrightarrow{\text{dogs catch a ball}} \mid \overrightarrow{\text{dogs catch a disease}}\rangle \quad =  \quad 0
\]
In an idealised case like this where there is very little (or no) overlap between the properties of the objects associated with one sense of ``catch'' (e.g. a disease), and those properties of the objects associated with another sense (e.g. a ball), disambiguation is perfect in that there is no similarity between the resulting phrases. In practice, in richer vector spaces, we would expect even diseases and balls to share some properties. However, as long as those shared properties are not those typically held by the object of catch, and as long as the usages of catch play to distinctive properties of diseases and balls, disambiguation will occur by the same mechanism as the idealised case above, and we can expect low similarity measures between such sentences.

\section{Related Work}
\label{sec:relatedwork}

Mitchell and Lapata introduce and evaluate a multiplicative model for
vector composition \cite{Lapata}. The
particular concrete construction of this paper differs from that
of~\cite{Lapata} in that our framework subsumes truth-theoretic as
well as corpus-based meaning, and our meaning construction relies on
and is guided by the grammatical structure of the sentence. The
approach of~\cite{ErkPado} is more in the spirit of ours, in that
extra information about syntax is used to compose meaning. Similar to
us, they use a structured vector space to integrate lexical
information with selectional preferences. Finally, Baroni and
Zamparelli model adjective-noun combinations by treating an adjective
 as a function from noun space to noun space, represented using a
 matrix, as we do in this paper \cite{BaroniEMNLP10}.


{\small
\bibliographystyle{plain}
\bibliography{iWCS}
}

\end{document}